\documentclass[letterpaper, 10pt, conference]{ieeeconf}
\IEEEoverridecommandlockouts
\usepackage{graphicx}
\usepackage{amsmath,amssymb}
\usepackage{booktabs}
\usepackage{tabularx}
\usepackage{xcolor}
\usepackage{xurl}
\usepackage[hidelinks]{hyperref}

\input{generated/keys.tex}

\title{\LARGE \bf The Latent That Never Was: A Forensic Re-run of the CVAE Ablation\\ in Action Chunking Transformers}

\author{Bo Kang%
\thanks{Bo Kang is with Ghent University, Belgium.}%
}

\begin{document}
\bstctlcite{icra:reference-style}
\maketitle
\thispagestyle{empty}
\pagestyle{empty}

\begin{abstract}
Action Chunking Transformers (ACT) are widely used to learn robot
manipulation from demonstrations. Their conditional variational autoencoder
includes an encoder meant to capture differences between demonstrations
during training. The original ACT paper reported that encoder removal
dropped the mean success rate from \F[0]{published.agg.vae}\% to
\F[0]{published.agg.novae}\% on two simulated tasks with human demonstrations.
We re-ran this ablation in the original code and checked whether the
findings depend on the implementation or training data. The published drop
does not reappear in our tests, although smaller gains or losses in success
rate remain uncertain. To investigate the discrepancy, we varied training length and
how checkpoints are selected for evaluation. Both can reverse which policy
scores higher, but the published drop's cause remains unknown.
Success rates alone leave open whether the encoder provides information
that helps the policy reconstruct demonstrated actions. On the tested ACT
benchmark, the sampled latent provides little reconstruction benefit at
every tested nonzero weight of the penalty on latent information.
At inference, ACT leaves this latent unused and sets it to zero.
Skipping the encoder increases training throughput in both implementations
we timed. We release code, evaluation tools and results so others can
repeat the comparisons and test the encoder on other tasks.\footnote{Experiments and results: \url{https://github.com/aida-ugent/act-cvae-forensics}}

\end{abstract}

\section{Introduction}\label{sec:intro}

\begin{figure}[t]
  \centering
  \includegraphics[width=\columnwidth]{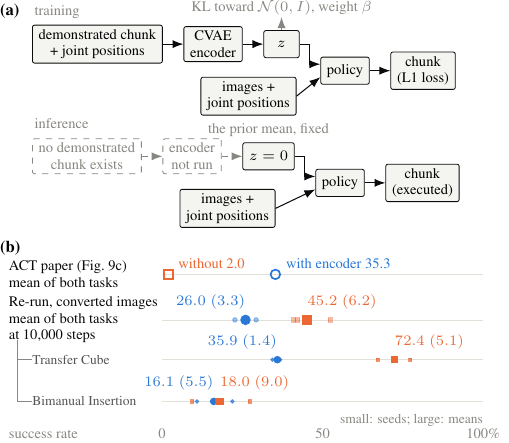}
  \caption{(a) ACT's encoder computes $z$ from the demonstrated chunk during
  training. At inference, ACT omits the encoder and uses zero.
  (b) The initial re-run on converted images reverses the
  published comparison, driven mainly by Transfer Cube
  (Sec.~\ref{sec:reversal}). Aggregate rows weight both tasks equally.
  Indented rows show the re-run's task components. Re-run labels give each
  arm's mean (SD) across \N{protocol.agg.10k.n_seeds} training seeds.
  Small marks show seeds and large filled marks show means. Hollow marks
  show published means, whose uncertainty is unreported. The re-run uses
  validation-selected checkpoints and temporal ensembling.}
  \label{fig:one}
\end{figure}

Action Chunking Transformers (ACT)~\cite{zhao2023act} are widely used
imitation-learning policies in robot manipulation. ACT uses a conditional
variational autoencoder (CVAE). Its encoder accounts for a third of the
parameters in the tested single-camera LeRobot model~\cite{cadene2024lerobot}. During training, the
encoder compresses each demonstrated action chunk, a short sequence of
actions, into a latent representation $z$. This latent is meant to capture
differences between demonstrations. At inference, ACT omits the encoder
and sets $z$ to zero, the prior mean (Fig.~\ref{fig:one}a).
The original ACT paper supports the encoder with an ablation on two
simulated tasks: handing over a cube in Transfer Cube and inserting a peg
into a socket in Bimanual Insertion. With human demonstrations, the mean
success rate dropped from
\F{published.agg.vae}\% to \F[0]{published.agg.novae}\% without the encoder
(Fig.~\ref{fig:one}b).

Yet that ablation has never been independently reproduced. The original
code has no removal switch, and the ablation reports neither training length,
seed counts nor uncertainty. The paper's main results use
\F{published.seeds} seeds and \F{published.rollouts} episodes per
seed~\cite[Table~\F{published.results.table}]{zhao2023act}; the code reuses
one fixed pose sample (Sec.~\ref{sec:resolution}).

Success rates alone cannot show whether the latent carries useful
information. ACT's decoder predicts actions from observations and may not
rely on $z$. Direct probes must test what the encoder learns and whether
it helps reconstruct demonstrated actions.

We therefore re-ran the ablation and probed the latent's role. We compare
the original code with LeRobot's ACT to test whether the encoder effect
depends on the implementation. Our compact adaptation of LeRobot's ACT, nanoACT,\footnote{\url{https://github.com/bokang-ugent/nanoACT}} makes
controlled changes and latent measurements easier to run. We vary training
length and how checkpoints are selected for evaluation to investigate the
published drop. We also vary the Kullback--Leibler (KL) penalty to test
whether it limits latent use (Sec.~\ref{sec:background}).

This paper answers three questions.
\begin{enumerate}
\item \textbf{Does the published drop reproduce?} The large drop does not
  reappear in our tested settings. The initial converted-image re-run
  reverses the comparison (Fig.~\ref{fig:one}b), while checks with original
  images leave smaller gains or losses uncertain
  (Sec.~\ref{sec:reproduction}).

\item \textbf{What could explain the published drop?} Training length and
  checkpoint selection can reverse which policy scores higher. However, this
  sensitivity does not establish the cause of the drop reported in the
  original ACT paper
  (Sec.~\ref{sec:forensics}).
\item \textbf{What does the latent carry?} The sampled latent provides little
  reconstruction benefit on ACT's tasks at the tested positive KL weights.
  On the planted-choice task, sufficiently small positive KL weights or
  removing the penalty can improve reconstruction, but our low-KL rollout
  test finds no success advantage (Sec.~\ref{sec:latent}).
  On PushT, a planar pushing task~\cite{chi2023diffusion}, the latent improves
  reconstruction at a KL weight below ACT's default (Sec.~\ref{sec:live}).
  These benefits alone do not establish improved task success,
  since ACT supplies zero at inference.
\end{enumerate}

Removing the encoder reduces training cost on ACT's benchmark, while the
measured success differences remain uncertain (Sec.~\ref{sec:discussion}).
All our experiments are in simulation.

\begin{table*}[t]
\caption{Implementation and control experiments grouped by research question.
Human-demonstration rows use converted images. Checks with original images are
reported in Secs.~\ref{sec:native} and \ref{sec:empty}.}
\label{tab:experiments}
\centering\small
\setlength{\tabcolsep}{\Fx{layout.experiment_map.padding}pt}
\begin{tabularx}{\textwidth}{@{}
>{\hsize=\Fx{layout.experiment_map.weight.experiment}\hsize\linewidth=\hsize\raggedright\arraybackslash\leavevmode}X
>{\hsize=\Fx{layout.experiment_map.weight.task}\hsize\linewidth=\hsize\raggedright\arraybackslash}X
>{\hsize=\Fx{layout.experiment_map.weight.training}\hsize\linewidth=\hsize\raggedright\arraybackslash}X
>{\hsize=\Fx{layout.experiment_map.weight.measurement}\hsize\linewidth=\hsize\raggedright\arraybackslash}X@{}}
\toprule
\textbf{Experiment} & \textbf{Task, data and code} & \textbf{Training setting} & \textbf{Comparison and measurement} \\
\midrule
\multicolumn{\Fx{layout.experiment_map.columns}}{@{}l@{}}{\textbf{R: Does the published drop reproduce?}} \\
\hypertarget{exp:rerun}{\F{layout.experiment_map.id.rerun}}~\F{published.budget.documented}-step re-run & Both ACT tasks; human demonstrations; original code
& \F{published.budget.documented} steps; three seeds
& Validation-loss selection with temporal ensembling; success rate on \F{protocol.te_poses} poses. \\
\hypertarget{exp:implementations}{\F{layout.experiment_map.id.implementations}}~Across implementations & Both ACT tasks: original and ours. LeRobot: Transfer Cube only. Human demonstrations.
& \F{protocol.budget.long} steps; three seeds
& Final checkpoints; success rate on \F{protocol.poses} poses. \\
\hypertarget{exp:scripted}{\F{layout.experiment_map.id.scripted}}~Scripted demonstrations & Both ACT tasks; demonstrations from a programmed controller; original and ours
& \F{protocol.budget.long} steps; three seeds. Original: \F{published.split.train}/\F{published.split.val} split; ours: all \F{published.demos} demonstrations.
& Final checkpoints; success rate on \F{protocol.poses} poses; \F{protocol.horizon.scripted}-step episodes for both tasks. \\
\midrule
\multicolumn{\Fx{layout.experiment_map.columns}}{@{}l@{}}{\textbf{E: What could explain the published drop?}} \\
\hypertarget{exp:training_length}{\F{layout.experiment_map.id.training_length}}~Training length & Transfer Cube; human demonstrations; original code
& Checkpoints at \F{protocol.budget.25k}, \F{protocol.budget.50k}, \F{protocol.budget.75k}, \F{protocol.budget.long} steps; three seeds
& Success rate on \F{protocol.poses} poses with chunked execution. Two curve seeds were retrained. \\
\hypertarget{exp:selection_execution}{\F{layout.experiment_map.id.selection_execution}}~Selection and execution & Both ACT tasks; original-code runs reused from the comparisons above
& \F{published.budget.documented} and \F{protocol.budget.long} steps; three seeds
& Final versus validation-selected checkpoints on \F{protocol.poses} poses; add ensembling to selected checkpoints only. \\
\hypertarget{exp:split_repeat}{\F{layout.experiment_map.id.split_repeat}}~Split repeat & Transfer Cube; human demonstrations; ours
& \F{published.budget.documented} steps; six seeds; original \F{published.split.train}/\F{published.split.val} split
& Final versus validation-selected checkpoints; success rate on \F{protocol.poses} poses. \\
\midrule
\multicolumn{\Fx{layout.experiment_map.columns}}{@{}l@{}}{\textbf{L: What does the latent carry?}} \\
\hypertarget{exp:latent}{\F{layout.experiment_map.id.latent}}~Latent measurements & Transfer Cube; human demonstrations; ours
& KL sweep at \F{protocol.budget.25k} steps; one seed. Default and zero-penalty runs also probed through \F{protocol.budget.long}.
& Latent information and reconstruction; predict demonstration-level motion properties without a KL penalty. \\
\hypertarget{exp:insertion}{\F{layout.experiment_map.id.insertion}}~Insertion check & Bimanual Insertion; human demonstrations; ours
& Default KL; three seeds at \F{protocol.budget.25k}, \F{protocol.budget.50k}, \F{protocol.budget.75k}, \F{protocol.budget.long} steps
& Check whether the latent findings extend to the other ACT task. \\
\hypertarget{exp:noise}{\F{layout.experiment_map.id.noise}}~Noise control & Transfer Cube; human demonstrations; ours
& \F{protocol.budget.long} steps; three seeds
& Replace encoder output with noise in training; success rate on \F{protocol.poses} poses. Reuses the with/without-encoder comparison above. \\
\hypertarget{exp:pusht}{\F{layout.experiment_map.id.pusht}}~PushT control & PushT demonstrations; ours
& \F{protocol.budget.25k} steps; one seed at each KL weight
& Check that latent probes detect informative~latents. \\
\hypertarget{exp:planted}{\F{layout.experiment_map.id.planted}}~Planted choice & Transfer Cube; \F{probe.coin.episodes} demonstrations from a programmed controller; ours
& \F{protocol.budget.25k} steps; one seed at each KL weight
& Test latent use when a choice is initially hidden from the observation. \\
\hypertarget{exp:images}{\F{layout.experiment_map.id.images}}~Image inputs & Transfer Cube; human and planted-choice demonstrations; ours
& \F{protocol.budget.25k} steps; two seeds per setting; KL weights depend on the data
& Test whether image inputs help the encoder supply useful information. \\
\hypertarget{exp:freebits}{\F{layout.experiment_map.id.freebits}}~Free bits & Transfer Cube; human demonstrations; ours
& \F{protocol.budget.25k} steps; one seed per allowance; default KL weight
& Vary the unpenalized KL allowance; measure latent information and reconstruction. \\
\bottomrule
\end{tabularx}
\par\smallskip
\raggedright\footnotesize
\textbf{Comparisons.} Success-rate comparisons use separately trained policies
with and without the encoder. Reconstruction probes compare errors with the
encoder's sampled latent and with zero, keeping the trained decoder fixed.
Section~\ref{sec:battery} defines the latent information and reconstruction measurements.
\par
\textbf{Training.} Seed counts are per configuration. Validation-selected
checkpoints may precede the stated training budget.
\par
\textbf{Evaluation.} Pose counts are per policy and seed. Temporal ensembling
averages overlapping predictions from chunks issued every step.
It uses \F{protocol.te_poses} poses, except for Transfer Cube at
\F{protocol.budget.long} steps, which uses \F{protocol.poses}.
\end{table*}

\section{Background and Related Work}\label{sec:related}

ACT can learn to predict actions without using its latent. We explain its
intended role, then review the unresolved success-rate evidence.

\subsection{ACT's Latent and Posterior Collapse}\label{sec:background}

ACT uses a CVAE to model differences between human
demonstrations~\cite{zhao2023act}. Its policy maps images and joint
positions to \F{published.chunk} successive actions, each specifying
\F{published.action_dim} joint targets.
ACT uses L1 loss, the mean absolute reconstruction error in
Eq.~\ref{eq:objective}. With one predicted chunk per observation,
minimizing this loss gives the median for each action element.
When demonstrations disagree, these medians can form a sequence nobody
demonstrated.

During training, the encoder reads the demonstrated chunk and joint
positions, without images. It predicts a Gaussian posterior over
\F{published.latent_dim} independent dimensions. The decoder receives a
sample $z$, images and joint positions. ACT calls $z$ a style variable
intended to represent differences between demonstrations. A separate $z$
is computed for each action chunk.

During training, the KL penalty limits information in $z$ by penalizing posterior
deviations from a fixed prior, the standard normal reference distribution
$\mathcal{N}(0,I)$. With penalty weight $\beta$, the training objective is
\begin{equation}\label{eq:objective}
\mathcal{L}=\underbrace{\operatorname*{mean}_{i}\,\lvert a_i-\hat a_i\rvert}_{\text{L1, a mean over the chunk}}
+\beta\underbrace{\sum_{d}\mathrm{KL}\big(q(z_d\mid a,o)\,\big\|\,\mathcal{N}(0,1)\big)}_{\text{KL, a sum over the latent}},
\end{equation}
where $a$ is the demonstrated chunk of \F{published.chunk_elements}
elements, normalized per joint coordinate, and $\hat a$ is its
reconstruction. The encoder sees joint positions $o$ and predicts posterior
$q$; $i$ and $d$ index action elements and latent dimensions.
The original CVAE formulation conditions its prior on the
observation~\cite{sohn2015cvae}; ACT uses the same prior for every
observation.

The original training command sets $\beta{=}\F{published.kl_weight}$.
One nat of summed KL then adds \F{published.kl_weight} to the loss,
compared with a mean reconstruction term near \F{probe.sweep.l1.hi} in
our trained models (Sec.~\ref{sec:empty}). Useful information can therefore increase total loss.
At inference, the future actions are unavailable; ACT fixes $z$ at zero.

ACT permits posterior collapse: the posterior matches the prior and the
decoder ignores $z$, predicting actions from the
observation~\cite{bowman2016,alemi2018elbo,lucas2019elbo}.
Training dynamics can also inhibit latent use~\cite{he2019lagging}.
Proposed remedies include gradually increasing KL weight~\cite{bowman2016},
unpenalized KL allowances (free bits)~\cite{kingma2016iaf}, and extra
encoder updates~\cite{he2019lagging}. These accounts motivate our probes
(Sec.~\ref{sec:why}).

\subsection{Prior Comparisons With and Without the Encoder}\label{sec:record}

Prior comparisons leave ACT's encoder effect unresolved. Its first author
wrote that the CVAE ``might not matter much''~\cite{zhao2023issue14}; the
code's tuning guide suggests a high KL weight or encoder
removal~\cite{act2024tuningtips}. Later systems and code releases omit the
encoder or make it optional without an ablation isolating its
effect~\cite{zhao2024unleashed,fu2024mobilealoha,fu2024humanplus}.
LeRobot enables it by default~\cite{cadene2024lerobot}.

Direct comparisons yield mixed results. SAM2Grasp~\cite{wu2025sam2grasp}
and two public notes~\cite{epicai2026notebook,colorfu1note} report an
encoder benefit on other tasks. Each note uses one training seed.
PAC-ACT favors removal when ACT is further trained with reinforcement
learning~\cite{pang2026pacact}.

A related study derives a KL-weight threshold for preserving distinct
behaviors, but does not train ACT~\cite{mazza2026multimodal}. Its
no-latent policy outperforms its fixed-prior CVAE on standard benchmarks,
with the ordering reversed on deliberately ambiguous tasks.
Failure to preserve behaviors does not
establish absence of latent information; ACT requires direct measurement.

We independently re-run ACT's CVAE ablation on its original tasks using
behavior cloning, report uncertainty across training seeds, test possible
explanations of the published drop, and directly measure latent use.

\section{Experimental Setup}\label{sec:setup}

A fair encoder comparison requires shared data, comparable implementations
and explicit training and evaluation choices. We establish these in turn,
then explain why latent probes are needed alongside success rates.
Table~\ref{tab:experiments} locates the experiments and their settings.

\subsection{Shared Data and Encoder Removal}\label{sec:fidelity}

In Transfer Cube, the robot moves a cube from its right gripper to its
left. In Bimanual Insertion, it inserts a peg held by its right gripper
into a socket held by its left. We score success at the original code's
final reward stage for each task.

Each task provides \F{published.demos} demonstrations from each of two
sources. Human demonstrations are collected by teleoperation;
scripted demonstrations are generated by a programmed controller.
The scripted data from the original ACT benchmark extend our comparison
beyond human demonstrations
(\hyperlink{exp:scripted}{\F{layout.experiment_map.id.scripted}}).

Our reference is the original implementation at revision
\texttt{\F{env.act.commit}}.\footnote{\url{https://github.com/tonyzhaozh/act}}
To use shared data across implementations, we recover demonstrations from
LeRobot's release. Joint positions and actions match the retrievable original
episodes exactly, but the images have undergone lossy video encoding.
We also repeat the original-code comparison with original ACT images
(Sec.~\ref{sec:native}).

To train without the encoder, we patch the original code.
The patched code omits the encoder, sets $z$ to zero during training, and
uses L1 loss alone. LeRobot's switch also omits the encoder; our implementation skips
its computation but retains its parameters.

For policies trained on human demonstrations, we retain the original
episode lengths: \F{published.horizon.tc} control steps on Transfer Cube
and \F{published.horizon.bi} on Bimanual Insertion.
Our evaluation script retains the original execution and scoring rules
while accepting externally supplied initial poses.
We use each checkpoint's normalization statistics to match the input
and action scaling in training.

\subsection{Implementation Checks}\label{sec:independence}

The implementation comparisons test whether the encoder effect depends on
the code. The three implementations form two code lineages because ours is adapted from
LeRobot's ACT~\cite{cadene2024lerobot}.

To check the simplified model's computations, we compared it with LeRobot's
ACT \F{env.lerobot} using identical weights, one batch and synchronized
random draws. Actions, posterior parameters and all parameters after one
optimizer step agreed exactly. A separate training check on scripted
Transfer Cube gave similar success rates with the same recipe and one seed
per implementation.

The original implementation uses MuJoCo \F{env.mujoco.original}; the other
two use MuJoCo \F{env.mujoco.gym} through gym-aloha. Because physics
versions differ, we compare encoder effects within each implementation
rather than absolute success rates across engines.

\subsection{Training and Evaluation Choices}\label{sec:poses}

The ACT paper does not specify the ablation's training length.
Our initial re-run
(\hyperlink{exp:rerun}{\F{layout.experiment_map.id.rerun}}) therefore uses the
\F{published.budget.documented}-step budget in the original code's README.
We retain the code's fixed
\F{published.split.train}/\F{published.split.val} training/validation
split and evaluate the checkpoint with the lowest validation loss,
which may precede the end of training.
The selected policy uses temporal ensembling: it predicts a new action
chunk every control step and averages overlapping predictions for that step.

To match training lengths, comparisons across implementations
(\hyperlink{exp:implementations}{\F{layout.experiment_map.id.implementations}}) and demonstration sources
(\hyperlink{exp:scripted}{\F{layout.experiment_map.id.scripted}})
use checkpoints saved after
\F{protocol.budget.long} steps.
This avoids validation-loss selection, whose criterion includes the KL
penalty only with the encoder.
The original code retains the
\F{published.split.train}/\F{published.split.val} split; LeRobot and ours
train on all \F{published.demos} demonstrations.
Within each implementation, both policies train on the same data.
These policies execute a full predicted action chunk before observing
again, rather than updating predictions every step.

We then vary training length, checkpoint selection and action execution
in the original code
(\hyperlink{exp:training_length}{\F{layout.experiment_map.id.training_length}},
\hyperlink{exp:selection_execution}{\F{layout.experiment_map.id.selection_execution}})
to test whether they explain the published drop.
In our implementation, a separate repeat
(\hyperlink{exp:split_repeat}{\F{layout.experiment_map.id.split_repeat}})
uses the original \F{published.split.train}/\F{published.split.val} split to
compare final and validation-selected checkpoints from the same runs.

Within each encoder comparison, both policies share initial poses to avoid
differences caused by separate evaluation samples.
This paired evaluation uses
\F{protocol.poses} poses per task for the implementation and scripted-data
comparisons, drawn once from the original sampler's ranges.
Temporal ensembling queries the policy every control step, increasing
evaluation cost. To limit this cost, some comparisons use fewer initial
poses (Table~\ref{tab:experiments}).

More poses reduce sampling uncertainty but cannot resolve variation between
training runs (Sec.~\ref{sec:resolution}). We therefore repeat training across
seeds, keeping the GPU model fixed within each seed's comparison. Each seed's
difference is the success rate with the encoder minus the rate without it.
The original-image check expands its training-seed sample to assess whether
its initial result depends on the sampled runs (Sec.~\ref{sec:native}).

\subsection{Latent Probes}\label{sec:pipeline}

Success rates alone cannot show whether the decoder uses latent information.
We therefore measure latent information and compare reconstruction with
sampled and zero latents in the same trained decoder. Positive controls
check that the probes detect useful latents (Sec.~\ref{sec:latent}).

\section{Does the Published Drop Reproduce?}\label{sec:reproduction}

The published drop does not reappear in our success-rate comparisons. After the initial
re-run, we check whether its outcome depends on the implementation,
demonstration source, images or training seeds. We then assess what the remaining
uncertainty permits.

\subsection{Re-run at \Fx{published.budget.documented} Steps}\label{sec:reversal}

Removing the encoder raises the mean success rate across both tasks in
our converted-image \F{published.budget.documented}-step re-run
(\hyperlink{exp:rerun}{\F{layout.experiment_map.id.rerun}}). It uses the original code,
validation-loss selection and temporal ensembling.
As in ACT's ablation, we average success rates equally across Transfer Cube
and Bimanual Insertion. The mean rises from \N{protocol.agg.10k.vae.mean}\% with the encoder to
\N{protocol.agg.10k.novae.mean}\% without it (Fig.~\ref{fig:one}b).
Even the lowest two-task mean without the encoder exceeds the highest
with it across all training seeds. Transfer Cube drives most of the
increase (Table~\ref{tab:rates},~\hyperlink{rates:tc-human-short}{e}--\hyperlink{rates:bi-human-short}{f}).

\subsection{Comparisons at Equal Training Lengths}\label{sec:matched}

Longer training on converted human-demonstration images does not reveal a large gain
in success rate from the encoder. The comparison across implementations
(\hyperlink{exp:implementations}{\F{layout.experiment_map.id.implementations}})
uses final checkpoints at \F{protocol.budget.long} steps, giving both
policies the same number of training updates.
For each task and implementation, the policies' mean success rates differ
by less than the standard deviation of the paired seed differences
(Table~\ref{tab:rates},~\hyperlink{rates:tc-human-long}{a}--\hyperlink{rates:bi-human-long}{b}).

We also test whether an encoder benefit appears with scripted demonstrations
(\hyperlink{exp:scripted}{\F{layout.experiment_map.id.scripted}})
instead of human demonstrations. There is no consistent encoder advantage
across the two tasks and implementations (Table~\ref{tab:rates},~\hyperlink{rates:tc-scripted-long}{c}--\hyperlink{rates:bi-scripted-long}{d}).
In the original code, removal
raises the mean Transfer Cube success rate by
\Nabs[1]{grid.original.tcs.100k.final.diff.mean} points,
with the same direction in every seed pair. On Bimanual Insertion, removal
lowers the mean by \Nabs[1]{grid.original.bis.100k.final.diff.mean} points,
but the effect changes direction between seeds.

\begin{table}[t]
\caption{Success rates with and without the encoder. Our human-demonstration
rows use converted images. Results with original images are in Sec.~\ref{sec:native}.}
\label{tab:rates}
\centering\footnotesize
\setlength{\tabcolsep}{\Fx{layout.rates.padding}pt}
\begin{tabular*}{\columnwidth}{@{\extracolsep{\fill}}llrrr@{}}
\toprule
\textbf{Code} & \textbf{Rule} & \textbf{With} & \textbf{Without} & \textbf{Diff.} \\
\midrule
\multicolumn{\Fx{layout.rates.columns}}{@{}l}{\textbf{\hypertarget{rates:tc-human-long}{(a)}~Transfer Cube, human, \F{protocol.budget.long} steps}} \\
original & final & \N{grid.original.tc.100k.final.vae.mean} (\N{grid.original.tc.100k.final.vae.sd}) & \N{grid.original.tc.100k.final.novae.mean} (\N{grid.original.tc.100k.final.novae.sd}) & $\N[1]{grid.original.tc.100k.final.diff.mean}$ (\N[1]{grid.original.tc.100k.final.diff.sd}) \\
original & val & \N{grid.original.tc.100k.val.vae.mean} (\N{grid.original.tc.100k.val.vae.sd}) & \N{grid.original.tc.100k.val.novae.mean} (\N{grid.original.tc.100k.val.novae.sd}) & $\N[1]{grid.original.tc.100k.val.diff.mean}$ \\
ours & final & \N{grid.nanoact.tc.100k.final.vae.mean} (\N{grid.nanoact.tc.100k.final.vae.sd}) & \N{grid.nanoact.tc.100k.final.novae.mean} (\N{grid.nanoact.tc.100k.final.novae.sd}) & $\N[1]{grid.nanoact.tc.100k.final.diff.mean}$ (\N[1]{grid.nanoact.tc.100k.final.diff.sd}) \\
ours, noise control (\hyperlink{exp:noise}{\F{layout.experiment_map.id.noise}}) & final & \multicolumn{2}{l}{\N{grid.nanoact.tc.100k.final.randz.mean} (\N{grid.nanoact.tc.100k.final.randz.sd}) with noise} & \\
LeRobot \F{env.lerobot} & final & \N{grid.lerobot.tc.100k.final.vae.mean} (\N{grid.lerobot.tc.100k.final.vae.sd}) & \N{grid.lerobot.tc.100k.final.novae.mean} (\N{grid.lerobot.tc.100k.final.novae.sd}) & $\N[1]{grid.lerobot.tc.100k.final.diff.mean}$ (\N[1]{grid.lerobot.tc.100k.final.diff.sd}) \\
\addlinespace[\Fx{layout.rates.group_gap}ex]
\multicolumn{\Fx{layout.rates.columns}}{@{}l}{\textbf{\hypertarget{rates:bi-human-long}{(b)}~Bimanual Insertion, human, \F{protocol.budget.long} steps}} \\
original & final & \N{grid.original.bi.100k.final.vae.mean} (\N{grid.original.bi.100k.final.vae.sd}) & \N{grid.original.bi.100k.final.novae.mean} (\N{grid.original.bi.100k.final.novae.sd}) & $\N[1]{grid.original.bi.100k.final.diff.mean}$ (\N[1]{grid.original.bi.100k.final.diff.sd}) \\
ours & final & \N{grid.nanoact.bi.100k.final.vae.mean} (\N{grid.nanoact.bi.100k.final.vae.sd}) & \N{grid.nanoact.bi.100k.final.novae.mean} (\N{grid.nanoact.bi.100k.final.novae.sd}) & $\N[1]{grid.nanoact.bi.100k.final.diff.mean}$ (\N[1]{grid.nanoact.bi.100k.final.diff.sd}) \\
\addlinespace[\Fx{layout.rates.group_gap}ex]
\multicolumn{\Fx{layout.rates.columns}}{@{}l}{\textbf{\hypertarget{rates:tc-scripted-long}{(c)}~Transfer Cube, scripted, \F{protocol.budget.long} steps}} \\
original & final & \N{grid.original.tcs.100k.final.vae.mean} (\N{grid.original.tcs.100k.final.vae.sd}) & \N{grid.original.tcs.100k.final.novae.mean} (\N{grid.original.tcs.100k.final.novae.sd}) & $\N[1]{grid.original.tcs.100k.final.diff.mean}$ (\N[1]{grid.original.tcs.100k.final.diff.sd}) \\
ours & final & \N{grid.nanoact.tcs.100k.final.vae.mean} (\N{grid.nanoact.tcs.100k.final.vae.sd}) & \N{grid.nanoact.tcs.100k.final.novae.mean} (\N{grid.nanoact.tcs.100k.final.novae.sd}) & $\N[1]{grid.nanoact.tcs.100k.final.diff.mean}$ (\N[1]{grid.nanoact.tcs.100k.final.diff.sd}) \\
\addlinespace[\Fx{layout.rates.group_gap}ex]
\multicolumn{\Fx{layout.rates.columns}}{@{}l}{\textbf{\hypertarget{rates:bi-scripted-long}{(d)}~Bimanual Insertion, scripted, \F{protocol.budget.long} steps}} \\
original & final & \N{grid.original.bis.100k.final.vae.mean} (\N{grid.original.bis.100k.final.vae.sd}) & \N{grid.original.bis.100k.final.novae.mean} (\N{grid.original.bis.100k.final.novae.sd}) & $\N[1]{grid.original.bis.100k.final.diff.mean}$ (\N[1]{grid.original.bis.100k.final.diff.sd}) \\
ours & final & \N{grid.nanoact.bis.100k.final.vae.mean} (\N{grid.nanoact.bis.100k.final.vae.sd}) & \N{grid.nanoact.bis.100k.final.novae.mean} (\N{grid.nanoact.bis.100k.final.novae.sd}) & $\N[1]{grid.nanoact.bis.100k.final.diff.mean}$ (\N[1]{grid.nanoact.bis.100k.final.diff.sd}) \\
\addlinespace[\Fx{layout.rates.group_gap}ex]
\multicolumn{\Fx{layout.rates.columns}}{@{}l}{\textbf{\hypertarget{rates:tc-human-short}{(e)}~Transfer Cube, human, \F{published.budget.documented} steps}} \\
original & final & \N{grid.original.tc.10k.final.vae.mean} (\N{grid.original.tc.10k.final.vae.sd}) & \N{grid.original.tc.10k.final.novae.mean} (\N{grid.original.tc.10k.final.novae.sd}) & $\N[1]{grid.original.tc.10k.final.diff.mean}$ \\
original & val & \N{grid.original.tc.10k.val.vae.mean} (\N{grid.original.tc.10k.val.vae.sd}) & \N{grid.original.tc.10k.val.novae.mean} (\N{grid.original.tc.10k.val.novae.sd}) & $\N[1]{grid.original.tc.10k.val.diff.mean}$ \\
original & val + TE & \N{grid.original.tc.10k.valte.vae.mean} (\N{grid.original.tc.10k.valte.vae.sd}) & \N{grid.original.tc.10k.valte.novae.mean} (\N{grid.original.tc.10k.valte.novae.sd}) & $\N[1]{grid.original.tc.10k.valte.diff.mean}$ \\
ours, split repeat (\hyperlink{exp:split_repeat}{\F{layout.experiment_map.id.split_repeat}}) & final & \N{grid.nanoact.tc.10kproto.final.vae.mean} (\N{grid.nanoact.tc.10kproto.final.vae.sd}) & \N{grid.nanoact.tc.10kproto.final.novae.mean} (\N{grid.nanoact.tc.10kproto.final.novae.sd}) & $\N[1]{grid.nanoact.tc.10kproto.final.diff.mean}$ \\
ours, split repeat (\hyperlink{exp:split_repeat}{\F{layout.experiment_map.id.split_repeat}}) & val & \N{grid.nanoact.tc.10kproto.val.vae.mean} (\N{grid.nanoact.tc.10kproto.val.vae.sd}) & \N{grid.nanoact.tc.10kproto.val.novae.mean} (\N{grid.nanoact.tc.10kproto.val.novae.sd}) & $\N[1]{grid.nanoact.tc.10kproto.val.diff.mean}$ \\
\addlinespace[\Fx{layout.rates.group_gap}ex]
\multicolumn{\Fx{layout.rates.columns}}{@{}l}{\textbf{\hypertarget{rates:bi-human-short}{(f)}~Bimanual Insertion, human, \F{published.budget.documented} steps}} \\
original & val + TE & \N{grid.original.bi.10k.valte.vae.mean} (\N{grid.original.bi.10k.valte.vae.sd}) & \N{grid.original.bi.10k.valte.novae.mean} (\N{grid.original.bi.10k.valte.novae.sd}) & $\N[1]{grid.original.bi.10k.valte.diff.mean}$ \\
\addlinespace[\Fx{layout.rates.group_gap}ex]
\multicolumn{\Fx{layout.rates.columns}}{@{}l}{\textbf{\hypertarget{rates:aggregate}{(g)}~Mean of both tasks, human}} \\
original, \F{protocol.budget.long} & val + TE & \N[1]{agg.vae.rate} & \N[1]{agg.novae.rate} & $\N[1]{agg.gap.true}$ \\
original, \F{published.budget.documented} (\hyperlink{rates:tc-human-short}{e},~\hyperlink{rates:bi-human-short}{f}) & val + TE & \N{protocol.agg.10k.vae.mean} (\N{protocol.agg.10k.vae.sd}) & \N{protocol.agg.10k.novae.mean} (\N{protocol.agg.10k.novae.sd}) & $\N{protocol.agg.10k.diff.mean}$ \\
published & & \F{published.agg.vae} & \F{published.agg.novae} & $+\F{published.agg.gap}$ \\
\bottomrule
\end{tabular*}
\par\smallskip
\raggedright
\textbf{Scores.} Our success rates are means across seeds (\%).
``Diff.'' is With minus Without (percentage points). Positive favors the encoder.
Parentheses give seed standard deviations, using paired differences for ``Diff.''
\par\smallskip
\textbf{Comparisons.} Final rows in (\hyperlink{rates:tc-human-long}{a}--\hyperlink{rates:bi-scripted-long}{d}) match training budgets;
val + TE in (\hyperlink{rates:tc-human-short}{e}--\hyperlink{rates:bi-human-short}{f}) uses released-code selection and execution in the re-run.
Other variants test selection, execution or noise substitution.
Group~(\hyperlink{rates:aggregate}{g}) averages tasks equally. The published ablation's training length and seed count are unreported~\cite{zhao2023act}.
\par\smallskip
\textbf{Settings.} Step counts are training budgets. ``final'' uses the end-of-budget checkpoint;
``val'' minimizes validation L1 plus weighted KL with the encoder, L1 alone without,
and can select an earlier step. TE adds temporal ensembling; our other rows execute full chunks.
\par\smallskip
\textbf{Our runs.} Three training seeds per configuration, six for the split repeat
(\hyperlink{exp:split_repeat}{\F{layout.experiment_map.id.split_repeat}}).
Policies share \F{protocol.poses} initial poses per seed.
TE uses \F{protocol.te_poses}, except Transfer Cube at \F{protocol.budget.long} steps uses \F{protocol.poses}.
Execution-only comparisons use common poses (Sec.~\ref{sec:knobs}).
\end{table}

\subsection{Checking the Input Images}\label{sec:native}

To check LeRobot's image conversion, we repeat the comparison with original
ACT images. Encoder removal no longer raises the mean success rate
(Sec.~\ref{sec:reversal}).
An initial \N{native.seed10k.historical.aggregate.n_seeds}-seed repeat in
the original code favored the encoder, but success-rate differences varied
widely across seed pairs. To improve precision, we added
\N{native.seed10k.fresh.aggregate.n_seeds} new seed pairs under the same
training and evaluation settings. At
\N{native.seed10k.steps} updates, mean success is
\Nabs{native.seed10k.pooled.aggregate.diff.mean} percentage points higher with
the encoder than without it across all \N{native.seed10k.pooled.aggregate.n_seeds}
seed pairs, weighting both tasks equally
(\F{protocol.ci}\% confidence interval:
$[\N{native.seed10k.pooled.aggregate.diff.ci_lo},
\N{native.seed10k.pooled.aggregate.diff.ci_hi}]$).
This uses validation-selected policies and temporal ensembling on
\N{native.seed10k.poses} shared poses.

To check the result after longer training, a separate
\N{native.budget.aggregate.100k.n_seeds}-seed original-image cohort uses final
\N{native.budget.steps.100k}-update checkpoints. Its equal-task mean difference is
\N{native.budget.aggregate.100k.diff.mean} points,
$[\N{native.budget.aggregate.100k.diff.ci_lo},
\N{native.budget.aggregate.100k.diff.ci_hi}]$, again using temporal ensembling
on shared poses.
Both intervals describe variation across paired training seeds and permit
higher or lower success after removal.

\subsection{Precision of the Comparisons}\label{sec:resolution}

Similar mean success rates do not establish equal performance with and
without the encoder.

\begin{samepage}
For original-code Transfer Cube on converted human demonstrations
(Table~\ref{tab:rates},~\hyperlink{rates:tc-human-long}{a}), the mean encoder advantage is
$\N[1]{grid.original.tc.100k.final.diff.mean}\pm\N{resolution.seeds3.t2df}$
percentage points (approximate \F{protocol.ci}\% confidence interval
across three training seeds). This uses final \F{protocol.budget.long}-step
checkpoints and \F{protocol.poses} shared poses. Both higher and lower
success after removal remain possible.
\par
\end{samepage}

The released evaluation script fixes its random seed and reuses the same
\F{published.rollouts} initial poses. New poses address evaluation-sampling
uncertainty, while independently trained policies address training-run
variation. Repeating the script adds neither.

\begin{figure}[t]
  \centering
  \includegraphics[width=\columnwidth]{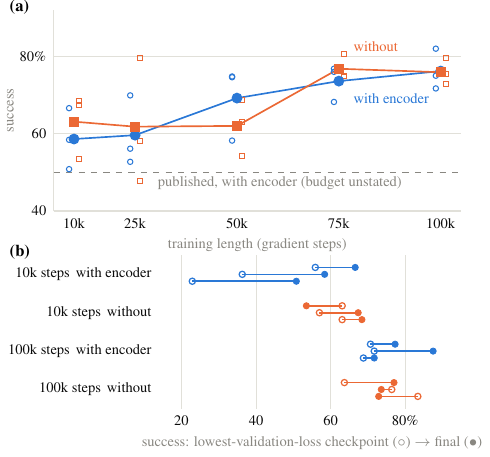}
  \caption{Training length and checkpoint selection can reverse which policy
  has the higher mean success rate. Both panels use original-code Transfer
  Cube with converted human-demonstration images. (a) Training length
  (\protect\hyperlink{exp:training_length}{\F{layout.experiment_map.id.training_length}}): hollow markers show individual seeds and filled markers
  their means. At the final point, two of the three seeds were retrained.
  The original-code final row uses the earlier runs
  (Table~\ref{tab:rates},~\protect\hyperlink{rates:tc-human-long}{a}).
  (b) Checkpoint selection
  (\protect\hyperlink{exp:selection_execution}{\F{layout.experiment_map.id.selection_execution}}): each segment joins one run's lowest-validation-loss checkpoint
  (hollow circle) to its final checkpoint (filled circle).}
  \label{fig:budget}
\end{figure}

\section{What Could Explain the Published Drop?}\label{sec:forensics}

Training and evaluation choices can reverse which policy scores higher,
but our tests do not establish the published drop's cause. We first ask
whether the encoder comparison depends on training length
(Sec.~\ref{sec:budget}), then on checkpoint selection and execution
(Sec.~\ref{sec:knobs}). We next examine differences in the physics engine
and data conversion (Sec.~\ref{sec:eliminated}), then identify the remaining
discrepancy (Sec.~\ref{sec:unresolved}).

\subsection{Training Length}\label{sec:budget}

The success-rate gap between policies with and without the encoder changes
with training length. The published ablation's training length is unknown,
so we cannot match its settings exactly.
The training-length test
(\hyperlink{exp:training_length}{\F{layout.experiment_map.id.training_length}}) extends the original-code
Transfer Cube runs with converted human-demonstration images from
\F{published.budget.documented} to \F{protocol.budget.long} steps.
We evaluate the checkpoints saved at each training length, using three seeds.

Both policies improve with training, while their gap changes sign
(Fig.~\ref{fig:budget}a). At every training length, the gap between the
policies' mean success rates is smaller than the standard deviation of the
paired seed differences.
A lead at one training length therefore does not establish a stable
encoder benefit.

\subsection{Checkpoint Selection and Action Execution}\label{sec:knobs}

\emph{Checkpoint selection.} Choosing which checkpoint to evaluate can reverse which policy has
the higher success rate within the same training runs. The original code selects by
validation loss. Validation loss combines reconstruction error and weighted
KL with the encoder, and uses reconstruction error alone without the encoder.
The policies therefore use different selection
criteria; neither directly measures rollout success.

The selection and execution test
(\hyperlink{exp:selection_execution}{\F{layout.experiment_map.id.selection_execution}})
compares final and validation-selected checkpoints from the same
original-code runs.
On Transfer Cube after \F{protocol.budget.long} training steps, the mean success
rate is \Nabs[1]{grid.original.tc.100k.final.diff.mean} points higher with the encoder
than without it at final checkpoints, but
\Nabs[1]{grid.original.tc.100k.val.diff.mean} points lower after validation selection
(Fig.~\ref{fig:budget}b).

At the \F{published.budget.documented}-step budget,
using selected rather than final checkpoints lowers the mean success rate with the encoder by
\Nabs[0]{selection.original.tc.10k.vae.mean} points, versus
\Nabs[0]{selection.original.tc.10k.novae.mean} without the encoder
(Table~\ref{tab:rates},~\hyperlink{rates:tc-human-short}{e}).

Selection also matters in our implementation, but here it enlarges an
encoder lead. The split repeat
(\hyperlink{exp:split_repeat}{\F{layout.experiment_map.id.split_repeat}})
uses Transfer Cube, the original split and six seeds at \F{published.budget.documented}
steps. Selection increases the mean encoder lead from
\N[1]{grid.nanoact.tc.10kproto.final.diff.mean} points at final checkpoints
to \N[1]{grid.nanoact.tc.10kproto.val.diff.mean} points
(Table~\ref{tab:rates},~\hyperlink{rates:tc-human-short}{e}). KL declines overall in these runs.
For the same runs with the encoder, selecting by L1 alone usually chooses an earlier
checkpoint than selecting by L1 plus weighted KL.

\emph{Execution.} ACT uses temporal ensembling at inference~\cite{zhao2023act}.
We test whether ensembling explains the discrepancy.
The execution comparison
(\hyperlink{exp:selection_execution}{\F{layout.experiment_map.id.selection_execution}})
uses checkpoints selected from original-code Transfer Cube runs trained
for \F{published.budget.documented} steps.
Keeping those checkpoints fixed, we compare ensembling with full-chunk execution on the same
\F{protocol.te_poses} poses.
Temporal ensembling raises the mean success rate without the encoder by
\Nabs[1]{te.original.tc.10k.common.novae.mean} points. With the encoder,
the mean falls by \Nabs[1]{te.original.tc.10k.common.vae.mean} points,
but the changes vary in direction across seeds. The policy with the encoder has a
lower mean success rate in both modes, so ensembling enlarges an existing
lead for encoder removal on Transfer Cube.

\subsection{Physics Engine and Data Conversion}\label{sec:eliminated}

Switching physics versions produces a small mean change in encoder advantage.
To check the version difference between implementations (Sec.~\ref{sec:independence}),
we evaluate the same final Transfer Cube policies from our implementation
in both engines. They were trained on human demonstrations for
\F{protocol.budget.long} steps with three seeds.
On \F{engine.scenes} shared initial poses, switching engines changes the
mean encoder advantage by
$\F{engine.interaction}\pm\F{engine.interaction.ci}$ percentage points
(\F{protocol.ci}\% confidence interval).

Joint positions and actions in LeRobot's release match the retrievable original
episodes exactly (Sec.~\ref{sec:fidelity}). The original-image encoder
comparison gives different success-rate estimates (Sec.~\ref{sec:native}).

\subsection{What Remains Unexplained}\label{sec:unresolved}

No tested setting reproduces the published
\F[0]{published.agg.novae}\% two-task mean without the encoder.
With checkpoint selection and temporal ensembling, our converted-image
two-task mean with the encoder is below the published mean at
\F{published.budget.documented} training steps and above it at
\F{protocol.budget.long} (Table~\ref{tab:rates},~\hyperlink{rates:aggregate}{g}). This ordering
does not establish the published ablation's training length or explain the low rate
without the encoder. Resolving that discrepancy would require the
records from the published ablation.

Success rates alone do not show what the encoder learns. We therefore
examine whether its latent carries information and whether that information
helps reconstruct demonstrated actions.

\section{What Does the Latent Carry?}\label{sec:latent}

On ACT's tasks, sampled latents give little reconstruction help at the
tested positive KL weights. We first define how to measure latent
information and use (Sec.~\ref{sec:battery}), then ask whether this small
reconstruction benefit persists across settings (Sec.~\ref{sec:empty}). Positive controls check
that the probes can detect useful latent information when it is present. Content probes
ask what information those latents carry (Sec.~\ref{sec:live}). Finally,
we test possible explanations for why sampled latents help reconstruction so little
(Sec.~\ref{sec:why}) and explain what limits latent use at
inference (Sec.~\ref{sec:boundary}).

\subsection{Measuring Latent Information and Use}\label{sec:battery}

The latent may carry information that the decoder already gets from the
observation. Here we define the measurements of latent information and
reconstruction benefit used in the following subsections.

\emph{Latent information.} We first track the posterior mean, the center
of the encoder's predicted distribution.
A latent dimension is an \emph{active unit} if this mean's variance across
demonstrated chunks and joint positions exceeds
\F{probe.active_threshold}~\cite{burda2016iwae}. The threshold can miss
smaller changes.

The decoder receives posterior samples, not posterior means. Sampling noise can make samples from
different chunks hard to distinguish, even when their posterior means differ.
We therefore also measure KL, summed over dimensions and averaged over inputs.
This gives an upper limit on the information samples carry about their
inputs, on average. Small KL means little information. Positive KL alone
does not establish information: the same distribution for every input can
still differ from the prior.

\emph{Reconstruction benefit.} The \emph{paired reconstruction test} compares
sampled and zero latents in the same trained decoder on identical batches.
Let $E_{\mathrm{zero}}$ be its mean absolute reconstruction error on
normalized actions with $z{=}0$, and $E_{\mathrm{sample}}$ the error with
posterior samples.
The \emph{$z$-advantage} is the relative error reduction:
\begin{equation}\label{eq:z-advantage}
z\text{-advantage}
=\frac{E_{\mathrm{zero}}-E_{\mathrm{sample}}}{E_{\mathrm{zero}}}
\times\F{math.percent_scale}\%.
\end{equation}
Positive values mean samples help; zero means no change, and negative
values mean samples hurt. In our implementation, repeated measurements vary
by a few tenths of a percentage point because of sampling and dropout.

\begin{figure}[t]
  \centering
  \includegraphics[width=\columnwidth]{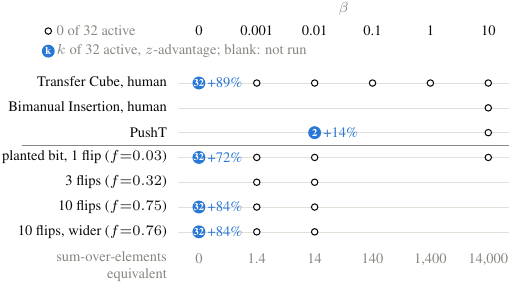}
  \caption{Reconstruction benefit is small at tested positive KL weights on ACT's tasks
  (\protect\hyperlink{exp:latent}{\F{layout.experiment_map.id.latent}},
  \protect\hyperlink{exp:insertion}{\F{layout.experiment_map.id.insertion}}),
  with positive controls on PushT
  (\protect\hyperlink{exp:pusht}{\F{layout.experiment_map.id.pusht}}) and without KL. The single-flip row is
  \protect\hyperlink{exp:planted}{\F{layout.experiment_map.id.planted}}.
  Other planted rows use more coin flips or larger motion offsets.
  Filled markers give active-unit counts and $z$-advantage: percentage
  reconstruction-error reduction relative to zero latents.
  Hollow markers indicate no active units (Sec.~\ref{sec:battery}). Blank cells were not run.
  Models use \F{protocol.budget.25k} training steps and one seed, except
  Transfer Cube at $\beta{=}0$ and Bimanual Insertion at
  \F{protocol.budget.long} steps. Insertion uses three seeds.
  In the planted rows, $f$ is the fraction of unpadded starts before the
  choice crosses a state-based visibility threshold. The lower axis gives
  equivalent KL weights for summed rather than averaged L1.}
  \label{fig:boundary}
\end{figure}

\subsection{KL Weight and Noise Substitution}\label{sec:empty}

At the default KL weight, sampled latents provide little reconstruction
benefit on Transfer Cube human demonstrations in our implementation. The latent measurements
(\hyperlink{exp:latent}{\F{layout.experiment_map.id.latent}}) at
$\beta{=}\F{published.kl_weight}$ find no active units and little benefit at
any probed checkpoint, from step \F{probe.first_step} to
\F{protocol.budget.long}.

Lower positive KL weights also give little reconstruction benefit.
The sweep trains one seed per weight for \F{protocol.budget.25k} steps,
with
$\beta\in\{\F{probe.beta.a},\F{probe.beta.b},\F{probe.beta.c},
\F{probe.beta.d},\F{published.kl_weight}\}$.
All have no active units and $z$-advantages near zero
(Eq.~\ref{eq:z-advantage}; Fig.~\ref{fig:boundary}), with reconstruction
error about \F{probe.sweep.l1.hi}. Even at the smallest weight, KL is only about
\F{probe.sweep.kl.b0001} nats, so sampled latents carry at most this much
input information on average.

The Insertion check
(\hyperlink{exp:insertion}{\F{layout.experiment_map.id.insertion}})
likewise finds no active units at the default weight across three seeds
per tested budget. At \F{protocol.budget.long} steps, $z$-advantages stay within
$\pm\F{probe.bi.zadv_max}\%$.

The original-code probes with original images also find little reconstruction
benefit at the default KL weight through \N{native.budget.steps.100k} updates. Across
\N{native.probe.models} runs with the encoder, replacing posterior samples with
zero or shuffling them between chunks changes reconstruction error little:
every validation comparison is within the predeclared
\N{native.probe.tolerance}\% relative-error tolerance. No latent unit is active,
and probes on training data give the same pattern. Small measurable
differences remain within this tolerance.

The reconstruction tests leave open whether encoder outputs affect task
success through training. The noise control
(\hyperlink{exp:noise}{\F{layout.experiment_map.id.noise}}) replaces these
outputs with standard normal noise during training.
Its encoder never runs. Inference still uses zero. For Transfer Cube human demonstrations,
policies trained with the encoder, with noise or without either have similar success rates
at final \F{protocol.budget.long}-step checkpoints (Table~\ref{tab:rates},~\hyperlink{rates:tc-human-long}{a}).
Their mean differences are smaller than their three-seed standard deviations.

\subsection{Positive Controls and Latent Content}\label{sec:live}

\emph{Positive controls.} To check that the small measured benefits do not
reflect insensitive probes, we test whether the same code and measurements
detect useful latents on PushT (\hyperlink{exp:pusht}{\F{layout.experiment_map.id.pusht}}),
the planar pushing task. At $\beta{=}\F{probe.beta.b}$,
\F{probe.pusht.active} of its \F{published.latent_dim} latent dimensions
are active: their posterior-mean variance across inputs exceeds the
threshold. Its $\F{probe.pusht.zadv}\%$ $z$-advantage means sampled
latents reduce reconstruction error relative to a zero latent in the same
trained decoder (Fig.~\ref{fig:boundary}). This confirms that the probe detects
reconstruction benefit on training batches.

To check for reconstruction benefit on an ACT task, we remove the KL
penalty on Transfer Cube human demonstrations
(\hyperlink{exp:latent}{\F{layout.experiment_map.id.latent}}).
The $z$-advantage reaches $\F{probe.b0.zadv.25k}\%$ at
\F{protocol.budget.25k} steps.
Together, these controls show that the probes detect reconstruction benefits
on PushT and on Transfer Cube without the KL penalty.

\emph{Latent content.} Better reconstruction does not establish that the
latent captures demonstration style, ACT's intended use.
We probe the Transfer Cube model trained without the KL penalty, where the
latent substantially improves reconstruction.
We ask whether posterior means predict timing, speed, smoothness and path
length of whole demonstrations. These \emph{style probes} learn from some
episodes and predict properties of other episodes. All episodes were used to train ACT.
We report the coefficient of determination ($R^2$). Zero means no improvement
over predicting the mean, and negative scores mean worse predictions.
The best test score at \F{protocol.budget.25k} steps is only
\F{probe.b0.proxy.best.25k}. More flexible, nonlinear probes at
\F{protocol.budget.long} steps also fail to predict these properties reliably.

To check whether the probes can recover other motion information, we instead
predict properties of the chunk the encoder sees, keeping the same episode
split. At \F{protocol.budget.25k} steps, scores reach
\F{probe.b0.chunk_speed} for mean speed and \F{probe.b0.chunk_jerk} for
mean squared jerk, where jerk is the rate of change of acceleration.
The probes recover chunk motion, but do not reliably recover whole-demonstration
timing, speed, smoothness or path length. Reconstruction benefit alone therefore
does not establish a style representation.

\subsection{Testing Reasons for the Small Reconstruction Benefit}\label{sec:why}

At the tested KL weights of \F{probe.beta.a} and above, changing the data
or adding encoder images gives little reconstruction benefit.

The planted-choice test
(\hyperlink{exp:planted}{\F{layout.experiment_map.id.planted}}) gives the encoder
information initially hidden from the decoder. We generate scripted Transfer
Cube demonstrations, choosing a high or low handover by coin flip before
scene generation. Every attempt succeeds, avoiding selection by outcome.
The encoder sees the demonstrated action chunk, which reveals the choice.
The decoder's observation initially does not, although later arm motion
reveals it. Reconstruction probes use chunks starting within the first
\F{probe.coin.early_window_end} steps, including starts after the choice becomes visible.
A hidden choice alone does not establish that its reconstruction benefit
would outweigh its KL cost.

At $\beta{=}\F{probe.beta.a}$, \F{probe.beta.b}, and
\F{published.kl_weight}, the planted-choice runs have no active units.
A separate classifier tests whether posterior means reveal which handover
height was chosen. We fit and test on different
episodes, all used to train ACT. It correctly predicts
\F{probe.coin.probe.b0001}--\F{probe.coin.probe.b001}\% of choices, versus
\F{probe.coin.chance}\% for random guessing. This suggests some choice
information in the means, yet samples supplied to the decoder give little
reconstruction benefit in the early window. Removing the KL
penalty raises classifier accuracy to
\F{probe.coin.probe.b0}\% and $z$-advantage to
$\F{probe.coin.b0.zadv}\%$ (Fig.~\ref{fig:boundary}).

ACT's encoder lacks the decoder's images, so it may miss which information
the decoder needs.
The image-input test
(\hyperlink{exp:images}{\F{layout.experiment_map.id.images}}) supplies pooled decoder image
features to the encoder, blocking encoder gradients into the image backbone.
On human and planted-choice data, whole-dataset $z$-advantages stay within
$\pm\F{imgenc.zadv_absmax}\%$ in the tested settings.

Finally, the free-bits test
(\hyperlink{exp:freebits}{\F{layout.experiment_map.id.freebits}})
allows some KL per dimension without penalty, permitting but not forcing
latent use.
At the default KL weight, larger allowances activate up to
\F{published.latent_dim} units, yet $z$-advantages remain near zero.
We report KL before the allowance.

KL cost, unequal encoder and decoder learning speeds, and sampling noise
remain possible contributors. Across three training seeds, the planted-choice
sweep yields early-window reconstruction
benefits of
\F{probe.smallkl.b1e4.min}--\F{probe.smallkl.b1e4.max}\% at
$\beta=\F{probe.smallkl.b1e4.beta}$ and
\F{probe.smallkl.b1e5.min}--\F{probe.smallkl.b1e5.max}\% at
$\beta=\F{probe.smallkl.b1e5.beta}$.
Yet at $\beta=\F{probe.smallkl.b1e5.beta}$, policies succeed less often
than matched no-encoder baselines.

\subsection{Latent Use at Inference}\label{sec:boundary}

Useful latents in reconstruction tests are unavailable during ACT's inference.
The encoder computes them from demonstrated action chunks, but ACT supplies
zero when choosing actions. In the planted-choice model trained without a KL penalty
(\hyperlink{exp:planted}{\F{layout.experiment_map.id.planted}}),
replacing the encoder's posterior sample with $z{=}0$ raises early-window
reconstruction error \F{probe.coin.b0.l1_ratio}-fold in the same trained decoder.

\section{Discussion}\label{sec:discussion}

Removing the encoder lowers training cost, but its effect on task success
remains uncertain in our experiments. We quantify the savings and discuss
the limits of our success-rate and latent findings.

\subsection{Training Cost of Encoder Removal}\label{sec:price}

Removing the encoder from the original code reduces stored parameters
from \N{native.params.vae_m} to \N{native.params.novae_m} million,
a reduction of \N{native.params.encoder_share}\%. LeRobot's removal switch also omits
the encoder, which accounts for \F{price.encoder_share_third}\% of the
tested single-camera model. These shares depend on the architecture.

On Transfer Cube human demonstrations with an RTX \F{price.ab.gpu},
bypassing encoder computation in our implementation raises training throughput by
\F{price.ab.throughput_gain}\% while retaining the encoder's parameters. Removing the
encoder from the original code raises throughput by
\F{price.original.timing.encoder_pct}\%.
We time one \F{price.ab.steps}-step run per policy in our implementation and three
\F{price.original.timing.steps}-step runs per policy in the original code.
These timing gains concern training: ACT already omits encoder computation
at inference. Unused encoder parameters could also be pruned from a policy
trained with the encoder.

\subsection{Limitations}\label{sec:limits}

The success-rate evidence is limited to the tested ACT implementations and
simulated tasks. Human and scripted demonstrations cover the same tasks.
The original-image seed extension and the separate longer-training check
remain too uncertain to establish whether encoder removal improves or
reduces success. Comparing policies on the same evaluation poses does not
remove variation caused by training with different random seeds.
The latent findings also need not extend to other tasks: on PushT, the
latent helps reconstruction.

The published ablation's unknown training length prevents an exact historical
match, and the published drop remains unexplained. The limited KL-weight
sweep also leaves the latent's training mechanism unresolved. The
planted-choice construction does not establish that encoding the choice
would lower total training loss.

\subsection{Conclusion}\label{sec:conclusion}

We do not reproduce the published success-rate drop in the tested settings,
and its cause remains unexplained. At the tested positive KL weights,
sampled latents provide little measured reconstruction benefit on this
benchmark. ACT uses zero instead at inference. Removing the encoder reduces
training cost, while its effect on task success remains uncertain.

We release the patches, probes, paired pose suites and results so others
can repeat the comparisons. More broadly, claims about a component's benefit
should specify the training and evaluation settings and be tested in
independent re-runs that account for variation across training seeds.

\section*{Acknowledgements}
The author thanks Nan Li for her careful review of the manuscript,
constructive suggestions, and helpful discussions.

The research leading to these results was co-funded by the European Union
(ERC, VIGILIA, 101142229), the Special Research Fund (BOF) of Ghent University
(BOF20/IBF/117), the Flemish Government under the
``Onderzoeksprogramma Artifici\"ele Intelligentie (AI) Vlaanderen'' programme,
and the FWO (project no. G073924N).

\bibliographystyle{IEEEtran}
\bibliography{refs}

\end{document}